\pdfoutput=1
\documentclass[11pt,a4paper]{article}
\usepackage[hyperref]{emnlp2020}
\usepackage{times}
\usepackage{latexsym}
\usepackage{graphicx}
\usepackage{CJKutf8}

\usepackage{microtype}

\aclfinalcopy  %Uncomment this line for the final submission
\def\aclpaperid{***}
\title{Modeling the Music Genre Perception across Language-Bound Cultures}

\author{\textbf{Elena V. Epure}\textsuperscript{\rm 1}, \textbf{Guillaume Salha}\textsuperscript{\rm 1,2}, \textbf{Manuel Moussallam}\textsuperscript{\rm 1}, \textbf{Romain Hennequin}\textsuperscript{\rm 1} \\
\textsuperscript{\rm 1} Deezer Research, Paris, France\\
\textsuperscript{\rm 2} LIX, \'{E}cole Polytechnique, Palaiseau, France \\ \texttt{research@deezer.com}}

\date{}

\begin{document}
\maketitle
\begin{abstract}
The music genre perception expressed through human annotations of artists or albums varies significantly across language-bound cultures.
These variations cannot be modeled as mere translations since we also need to account for cultural differences in the music genre perception.
In this work, we study the feasibility of obtaining relevant cross-lingual, culture-specific music genre annotations based only on language-specific semantic representations, namely distributed concept embeddings and ontologies. 
Our study, focused on six languages, shows that unsupervised cross-lingual music genre annotation is feasible with high accuracy, especially when combining both types of representations.
This approach of studying music genres is the most extensive to date and has many implications in musicology and music information retrieval.
Besides, we introduce a new, domain-dependent cross-lingual corpus to benchmark state of the art multilingual pre-trained embedding models. 
\end{abstract}

\section{Introduction}
\label{sec:intro}
A prevalent approach to culturally study music genres starts with a common set of music items, \emph{e.g.} artists, albums, tracks, and assumes that the same music genres would be associated with the items in all cultures \cite{Ferwerda2016InvestigatingTR,skowron2017predicting}.
However, music genres are subjective.
Cultures themselves and individual musicological backgrounds influence the music genre perception, which can differ among individuals \cite{Sordo2008,Lee2013KPopGA}.  
For instance, a Westerner may relate \emph{funk} to \emph{soul} and \emph{jazz}, while a Brazilian to \emph{baile funk} that is a type of \textit{rap} \cite{hennequinaudio}.
Thus, accounting for cultural differences in music genres' perception could give a more grounded basis for such cultural studies. 
However, ensuring both a common set of music items and culture-sensitive annotations with broad coverage of music genres is strenuous
\cite{bogdanov2019acousticbrainz}.

To address this challenge, we study the feasibility of cross-culturally annotating music items with music genres, without relying on a parallel corpus.
In this work, culture is related to a community speaking the same language \cite{kramsch1998language}.
The specific research question we build upon is: 
assuming consistent patterns of music genres association with music items within cultures, can a mapping between these patterns be learned by relying on language-specific semantic representations?
It is worth noting that, since music genres fall within the class of Culture-Specific Items \cite{Aixela1996,Newmark1988}, cross-lingual annotation, in this case, cannot be framed as standard translation, as one also needs to model the dissimilar perception of music genres across cultures.

Our work focuses on four language families, Germanic (English-\textbf{en} and Dutch-\textbf{nl}), Romance (Spanish-\textbf{es} and French-\textbf{fr}), Japonic (Japanese-\textbf{ja}), Slavic (Czech-\textbf{cs}), and on two types of language-specific semantic representations, ontologies and multi-word expression embeddings.

First, ontologies are often used to represent music genres, showing how they relate conceptually \cite{schreiber2016genre}.
We identify Wikipedia\footnote{\url{https://en.wikipedia.org}}, the online multilingual encyclopedia, to be particularly relevant to our study.
It extensively documents worldwide music genres relating them through a coherent set of relation types across languages (\emph{e.g.} \emph{derivative genre}, \emph{sub-genre}).
Though the relations types are the same per language, the actual music genres and the way they are related can differ.
Indeed, Pfeil et al. \citep{Pfeil2006} have shown that Wikipedia contributions expose cultural differences aligned with the ones in the physical world.

Second, music genres can be represented from a distributional semantics perspective. 
Word vector spaces are generated from large corpora following the distributional hypothesis, \emph{i.e.} words with similar contexts have akin meanings.
As languages are passed on culturally, we assume that the language-specific corpora used to create these spaces are sufficient to convey concepts' cultural specificity into their vector representations.
In our study, we focus on multiple recent multilingual pre-trained models to generate word or sentence\footnote{Music genres can be multi-word expressions.
Previous work \cite{Shwartz2019} successfully embed phrases with sentence embedding models.
In particular, contextualized language models result in more meaningful representations for diverse composition tasks.} embeddings \cite{arora2016simple,grave-etal-2018-learning,Artetxe-Schwenk-2019,devlin-etal-2019-bert,lample2019cross}, to account for variances in the used corpora or model designs.

Lastly, we combine the semantic representations by retrofitting distributed music genre embeddings to music genre ontologies.
Retrofitting \cite{faruqui-etal-2015-retrofitting} modifies each concept embedding such that the representation is still close to the distributed one, but also encodes ontology information. 
Initially, we retrofit music genres per language, using monolingual ontologies.
Then, by partially aligning these ontologies, we apply retrofitting to learn multilingual embeddings from scratch.

The results show that we can model the cross-lingual music genre annotation with high accuracy by combining both types of language-specific semantic representations.
When comparing the representations derived from multilingual pre-trained models, the smooth inverse frequency averaging \cite{arora2016simple} of aligned word embeddings outperforms the state of the art approaches.
To our knowledge, this simple method has been rarely used to embed multilingual sentences \cite{vargas-etal-2019-multilingual}, and we hypothesize its potential as a strong baseline on other cross-lingual datasets and tasks too.
Finally, embedding learning based on retrofitting leads to better multilingual music genre representations than when inferred with pre-trained embedding models.
This opens the possibility to learn embeddings for rare music genres or languages when aligned music genre ontologies are available.

Summing up, our contributions are: 
1) a study on how effective language-specific semantic representations of music genres are for modeling cross-lingual annotation, without relying on a parallel music item corpus; 
2) an extensive evaluation of multilingual pre-trained embedding models to derive representations for multi-word concepts in the music domain.
Our study can enable complete musicological research, but also localized music information retrieval. 
This latter application is crucial for online music streaming platforms that leverage music genre annotations to provide worldwide, user-personalized music recommendations.

Our domain-specific study complements other works benchmarking general-language sentence representations \cite{conneau-etal-2018-xnli}.
Finally, we provide an in-depth formal analysis of the retrofitting part of our method. 
We prove the strict convexity of retrofitting and show that the ontology concepts' final embeddings converge to the same values despite the order in which concepts are iteratively updated, on condition that we know a single initial node embedding in each connected component of the ontology.

\section{Related Work}
\label{sec:background}
Music genres are conceptual representations encompassing a set of conventions between the music industry, artists, and listeners about individual music styles  \cite{lena2012banding}.
From a cultural perspective, it has been shown that there are differences in how people listen to music genres. \cite{Ferwerda2016InvestigatingTR,skowron2017predicting}.
Average listening habits in some countries span across many music genres and are less diverse in other countries \cite{Ferwerda2016InvestigatingTR}. 
Also, cultural dimensions proved strong predictors for the popularity of specific music genres \cite{skowron2017predicting}.

Despite the apparent agreement on the music style for which the music genres stand, conveyed in the earlier definition and implied in the related works too, music genres are subjective concepts \cite{Sordo2008,Lee2013KPopGA}.
To address this subjectivity, \citet{bogdanov2019acousticbrainz} proposed a dataset of music items annotated with English music genres by different sources.
In this line of work, we address the divergent perception of music genres. Still, we focus on multilingual, unsupervised music genre annotation without relying on content features, \emph{i.e.} audio or lyrics.
We also complement similar studies in other domains (art: \citealp{Eleta2012astudy}) with another research method.

Then, in the literature, there are other works that benchmark pre-trained word and sentence embedding models.
\citet{heijden2019comparison} compares multilingual contextual language models for named-entity recognition and part-of-speech tagging.
\citet{shwartz-dagan-2019-still} use multiple static and contextual word embeddings to represent multi-word expressions and assess their capacity to capture meaning shift and implicit meaning in compositionality.
\citet{conneau-etal-2018-xnli} formulate a new task to evaluate cross-lingual sentence representation centered on natural language inference.

Compared to these works, our benchmark is aimed at cross-lingual annotation;
we target a specific domain, music, for which we try concept embedding adaptation with retrofitting;
and we also test a multilingual sentence representation obtained with smooth inverse frequency averaging of multilingual word embeddings.
As discussed in a recent survey on cross-lingual word embedding models \cite{Ruder2019}, there is a need to unlock domain-specific data to assess if general-language sentence representations are also accurate across domains.
Our work builds towards this goal.

\section{Cross-lingual Music Genre Annotation}

\begin{CJK}{UTF8}{ipxm}
\begin{figure*}[t]
\centering
\includegraphics[width=0.65\textwidth]{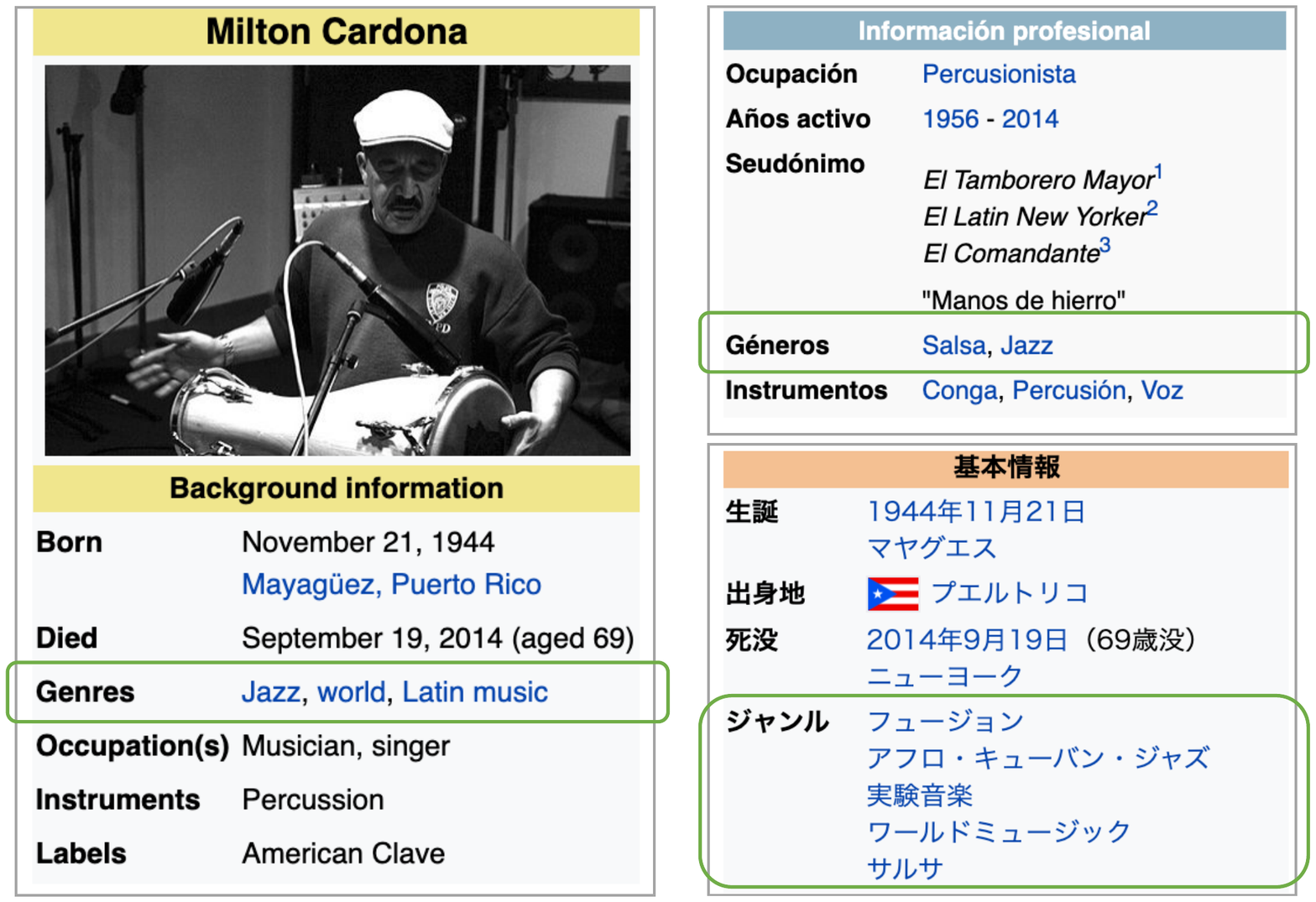}
\caption{Wikipedia infoboxes of Puerto Rican artist \textit{Milton Cardona}, in English (\textbf{en}), Spanish (\textbf{es}) and Japanese (\textbf{ja}) languages. Some music genre annotations are culture-specific, such as \textit{World}/ワールドミュージック which is present for \textbf{en} and \textbf{ja} but not for \textbf{es}, or \textit{実験音楽} (\textit{Experimental Music}) for \textbf{ja} only.}\label{fig:example}
\end{figure*}
\end{CJK}

\label{sec:crossannotation}
Further, we formalize the cross-lingual annotation task and the strategy to evaluate it in Section \ref{sec:problem}.
We describe the test corpus used in this work, together with its collection procedure in Section \ref{sec:testcorpus}.

\subsection{Problem Formalization}
\label{sec:problem}
The cross-lingual music genre annotation consists of inferring, for music items, tags in a target language $L_t$, knowing tags in a source language $L_s$.
For instance, knowing the English music genres of \emph{Fatboy Slim} (\emph{big beat}, \emph{electronica}, \emph{alternative rock}), the goal is to predict \emph{rave} and \emph{rock alternativo} in Spanish.
As shown in the example, but also Section \ref{sec:intro}, the problem goes beyond translation and instead targets a model able to map concepts, potentially dissimilar, across languages and cultures.

Formally, given $S$ a set of tags in language $L_s$, $\mathcal{P}$ the partitions of $S$ and $T$ a set of tags in language $L_t$, a mapping scoring function $f: \mathcal{P}(S) \rightarrow {\rm I\!R}^{|T|}$ can attribute a prediction score to each target tag, relying on subsets of source tags drawn from $S$ \cite{hennequinaudio,Epure2019,Epure2020}.
The produced score incorporates the degree of relatedness of each particular input source tag to the target tag.
A common approach to compute relatedness in distributional semantics relies on cosine similarity.
Thus, for $\{ s_{1}, ..., s_K \}$ source tags and any target tag $t$, $f$ can be defined as:
\begin{equation}
f_t(\{s_{1}, s_{2}, \dots, s_K\}) = \frac{1}{K} \sum_{k=1}^{K}\frac{{\textbf{s}_{k}}^T\textbf{t}}{||\textbf{s}||_2 ||\textbf{t}||_2},
\label{eq:transf}
\end{equation} 
where $|| \cdot ||_2$ is the Euclidean norm.

\subsection{Test Corpus}
\label{sec:testcorpus}
Wikipedia records worldwide music artists and their discographies, with a frequent mentioning of their music genres.
By manually checking the Wikipedia pages of miscellaneous music items, we observed that their music genres vary significantly across languages.
For instance, \textit{Knights of Cydonia}, a single by \textit{Muse}, was annotated in Spanish as \textit{progressive rock}, while in Dutch as \textit{progressive metal} and \textit{alternative rock}.
In Figure \ref{fig:example}, we show another example of different annotations in English, Spanish, and Japanese from Wikipedia infoboxes.
As Wikipedia writing is localized, contributors' culture can lead to differences in the multilingual content on the same topic \cite{Pfeil2006}, particularly for subjective matters.
Thus, Wikipedia was a suitable source for assembling the test corpus.

\begin{table}[t]
\begin{small}
\centering
\begin{tabular}{|l|lllll|}
\hline
\textbf{Language}  & \textbf{nl} & \textbf{fr} & \textbf{es} & \textbf{cs} & \textbf{ja} \\
\hline
\textbf{en} &  12604  & 28252 & 32891 & 4772 & 14752 \\
\textbf{nl} & & 7139 & 7689 & 1885 & 3426 \\
\textbf{fr} & & &  15616 & 3046 & 8622 \\
\textbf{es} & & & & 3245 & 7644 \\
\textbf{cs} & & & & & 2065 \\
\hline
\end{tabular}
\caption{\label{tab:corpus_items} Number of music items for language pair. }
\end{small}
\end{table}

\begin{table}[t]
\begin{small}
\centering
\begin{tabular}{|l|l|l|}
\hline
 & \multicolumn{2}{c|}{\textbf{Number of unique music genres}}  \\ 
\textbf{Language} & \textbf{Corpus} \textit{(Avg. per item)} & \textbf{Ontology} \\ \hline
\textbf{en} & 558 \textit{(2.12 $\pm$ 1.34)} & 10748 \\
\textbf{nl} & 204 \textit{(1.71 $\pm$ 1.06)} & 1529 \\
\textbf{fr} & 364 \textit{(1.75 $\pm$ 1.06)} & 2905 \\
\textbf{es} & 525 \textit{(2.11 $\pm$ 1.34)} & 3988\\
\textbf{cs} & 133 \textit{(2.23 $\pm$ 1.34)} & 1418\\
\textbf{ja} & 192 \textit{(1.51 $\pm$ 1.11)} & 1609\\
\hline
\end{tabular}
\caption{\label{tab:no_genres} Number of unique music genres in the corpus (Section \ref{sec:testcorpus}) and in the ontology (Section \ref{sec:ontology}).}
\end{small}
\end{table}

Using DBpedia \cite{Auer_2007} as a proxy to Wikipedia, we collected music items such as artists and albums, annotated with music genres in at least two of the six languages (\textbf{en}, \textbf{nl}, \textbf{fr}, \textbf{es}, \textbf{cs} and \textbf{ja}).
We targeted \emph{MusicalWork}, \emph{MusicalArtist} and \emph{Band} DBpedia resource types, and we only kept music items that were annotated with music genres which appeared at least $15$ times in the corpus.
Our final corpus includes 63246 music items.
The number of annotations for each language pair is presented in Table \ref{tab:corpus_items}.
We also show in Table \ref{tab:no_genres} the number of unique music genres per language in the corpus and the average number of tags for each music item.

The \textbf{en} and \textbf{es} languages use the most diverse tags.
This can be because more annotations exist in these languages, in comparison to \textbf{cs}, which has the least annotations and least diverse tags.
However, the mean number of tags per item appears relatively high for \textbf{cs}, while \textbf{ja} has the smallest mean number of tags per item.

\section{Language-specific Semantic Representations for Music Genres}

\label{sec:resources}
This work aims to assess the possibility of obtaining relevant cross-lingual music genre annotations, able to capture cultural differences too, by relying on language-specific semantic representations.
Two types of semantic representations are investigated given their popularity: ontologies to represent music genre relations (presented in Section \ref{sec:ontology}) and distributed embeddings to represent multi-word expressions in general (presented in Section \ref{sec:embs}).
In contrast to this unsupervised approach, mapping patterns of associating music genres with music items across cultures could have also been enabled with a parallel corpus.
However, gathering a corpus that includes all music genres for each pair of languages is challenging.

\subsection{Music Genre Ontology}
\label{sec:ontology}
Conceptually, music genres are interconnected entities.
For example, \emph{rap west coast} is a sub-genre of \emph{hiphop} or \emph{música electrónica} is the origin of \emph{synthpunk}.
Academic and practitioner communities often use ontologies or knowledge graphs to represent music genre relations and enrich the music genre definitions \cite{schreiber2016genre,Lisena2018}.
As  mentioned in Section \ref{sec:intro}, we use in this study Wikipedia-based music genre ontologies because the multilingual Wikipedia contributions on the same topic can differ and these differences have been proven aligned with the ones in the physical world \cite{Pfeil2006}.

We further describe how we crawl the Wikipedia-based music genres ontologies for the six languages by relying on DBpedia.
For each language, first, we constitute the seed list using two sources: the DBpedia resources of type \textit{MusicGenre} and their aliases linked through the \textit{wikiPageRedirects} relation; the music genres discovered when collecting the test corpus (introduced in Section \ref{sec:problem}) and their aliases.
Then, music genres are fetched by visiting the DBpedia resources linked to the seeds through the relations \textit{wikiPageRedirects}, \textit{musicSubgenre}, \textit{stylisticOrigin}, \textit{musicFusionGenre} and \textit{derivative}\footnote{We present these relations by their English names, which may be translated in the DBpedia versions in other languages.}.
The seed list is updated each time, allowing the crawling to continue until no new resource is found.

In DBpedia, resources are sometimes linked to their equivalents in other languages through the relation \textit{sameAs}.
For most experiments, we rely on monolingual music genres ontologies. 
However, we also collect the cross-lingual links between music genres to include a translation baseline for cross-lingual annotation, \emph{i.e.} for each music genre in a source language, we predict its equivalent in a target language using DBpedia.
Besides, we try to learn aligned embeddings from scratch by relying on these partially aligned music genre ontologies, as will be discussed in Section \ref{sec:retro}.

The number of unique Wikipedia music genres discovered in each language is presented in Table~\ref{tab:no_genres}.
Let us note that the graph numbers are much larger than the test corpus numbers, emphasizing the challenge to constitute a parallel corpus that covers all language-specific music genres.

\subsection{Music Genre Distributed Representations}
\label{sec:embs}
As music genres are multi-word expressions, we make use of existing sentence representation models.
We also inquire into word vector spaces and obtain sentence embeddings by hypothesizing that music genres are generally compositional, \emph{i.e.} the sense of a multi-word expression is conveyed by the sense of each composing word (\emph{e.g.} \textit{West Coast rap}, \textit{jazz blues}; there are also non-compositional examples like \textit{hard rock}).
We set our investigation scope to multilingual pre-trained embedding models, and we consider both static and contextual word/sentence representations as described next.

\paragraph{Multilingual Static Word Embeddings.} The classical word embeddings we study are the multilingual fastText word vectors trained on Wikipedia and Common Crawl \cite{grave-etal-2018-learning}. 
The model is an extension of the Common Bag of Word Model (CBOW, \citealp{Mikolov2013}), which includes subword and word position information.
The fastText word vectors are trained in distinct languages.
Thus, we must ensure that the monolingual word vectors are projected in the same space for cross-lingual annotation.
We perform the alignment with the method proposed by \citet{joulin-etal-2018-loss}, which treats word translation as a retrieval task and introduces a new loss relying on a relaxed cross-domain similarity local scaling criterion.

\paragraph{Multilingual Contextual Word Embeddings}.
Contextual word embeddings \cite{peters-etal-2017-semi,devlin-etal-2019-bert}, in contrast to the classical ones, are dynamically inferred based on the given context sentence. 
This type of embedding can address polysemy as the word sense is disambiguated through the surrounding text.
In our work, we include two recent contextualized language models compatible with the multilingual scope: multilingual Bidirectional Encoder Representations from Transformers (BERT, \citealp{devlin-etal-2019-bert}) and Cross-lingual Language Model (XLM, \citealp{lample2019cross}).

BERT \cite{devlin-etal-2019-bert} is trained to jointly predict a masked word in a sentence and whether sentences are successive text segments.
Similar to fastText \cite{grave-etal-2018-learning}, subword and word position information is also used.
An input sentence is tokenized against a limited token vocabulary with a modified version of the byte pair encoding algorithm (BPE, \citealp{sennrich-etal-2016-neural}).
Multilingual BERT is trained as a single-language model, fed with 104 concatenated monolingual Wikipedias \cite{pires2019multilingual,wu-dredze-2019-beto}.  

XLM \cite{lample2019cross} has a similar architecture to BERT.
Also, it shares with BERT one training objective, the masked word prediction, and the tokenization using BPE, but applied on sentences differently sampled from each monolingual Common Crawl corpus.
Compared to BERT, two other objectives are introduced, to predict a word from previous words and a masked word by leveraging two parallel sentences.
Thus, to train XLM, several multilingual aligned corpora are used \cite{lample2019cross}.

\paragraph{Multilingual Sentence Embeddings.}
Contextualized language models can be exploited in multiple ways.
First, as \citet{lample2019cross} show, by training the transformers on multi-lingual data, cross-lingual word vectors are obtained in an unsupervised way.
The word vectors can be accessed through the model lookup table.
These embeddings are merely aligned but not contextual, thus directly comparable to fastText.
For these three types of cross-lingual non-contextual word embeddings, fastText (\textbf{FT}), the multilingual BERT's lookup table (\textbf{mBERT}) and the XLM's lookup table (\textbf{XLM}), we compute the sentence embedding using the standard average (\textit{avg}) or the smooth inverse frequency averaging (\textit{sif}) introduced by \citet{arora2016simple}. 

Formally, let $c$ denote a music genre composed of multiple tokens $\{t_1, t_2, \dots, t_M\}$, $\textbf{t}_m$ the embedding of each token $t_m$ initialized from a given pre-trained embedding model or to $d$-dimensional\footnote{In Appendix \ref{app:results}, we report $d$ for each embedding type.} null vector $\textbf{0}_d$ if $t_m$ is absent from the model vocabulary, and $\hat{\textbf{q}}_i \in {\rm I\!R}^d$ the representation of $c$ which we want to infer.
The \textit{avg} strategy computes $\hat{\textbf{q}}_i$ as
$\frac{1}{M}\sum_{m=1}^M{\textbf{t}_m}$.
The \textit{sif} strategy computes $\hat{\textbf{q}}_i$ as:
\begin{equation}
  \overline{\textbf{q}}_i = \frac{1}{M}\sum_{m=1}^M{\frac{a}{a + f_{t_{m}}}\textbf{t}_{m}}  
\label{eq:sif1}
\end{equation}
\begin{equation}
    \hat{\textbf{q}}_i = \overline{\textbf{q}}_i - \textbf{u}\textbf{u}^T\overline{\textbf{q}}_i
\end{equation}
where $f_{t_{m}}$ is the frequency of $t_m$, $a$ is a hyper-parameter usually fixed to $10^{-3}$ \cite{arora2016simple} and $\textbf{u}$ is the first singular vector obtained through the singular value decomposition \cite{Golub:1970} of $\overline{\textbf{Q}}$, the embedding matrix computed with the Equation \ref{eq:sif1} for all music genres.
Vocabulary tokens of pre-trained embedding models are usually sorted by decreasing frequency in the training corpus, \emph{i.e.} the higher the rank, the more frequent the token.
Thus, based on the Zipf's law \cite{Zipf1949}, $f_{w_m}$ can be approximated by $1/z_{t_{m}}$, $z_{t_{m}}$ being the rank of $t_m$.
The intuition of this simple sentence embedding method is that uncommon words are semantically more informative.

Second, contextualized language models can be used as feature extractors representing sentences from the contextual embeddings of the associated tokens.
Multiple strategies exist to retrieve contextual token embeddings: to use the embeddings layer or the last hidden layer or to apply min or max pooling over time \cite{devlin-etal-2019-bert}. 
To infer a fixed-length representation of a multi-word music genre, we try max and mean pooling over token embeddings \cite{lample2019cross,reimers-gurevych-2019-sentence}, obtained with the diverse strategies mentioned before. 
We denote these sentence embeddings $\textbf{XLM}_{Ctxt}$ and $\textbf{mBERT}_{Ctxt}$.

The contextualized language models can be further fine-tuned for particular downstream tasks, yielding better sentence representations \cite{eisenschlos-etal-2019-multifit,lample2019cross}.
Existing evaluations of cross-lingual sentence representations are centered on natural language inference (XNLI, \citealp{conneau-etal-2018-xnli}) or classification \cite{eisenschlos-etal-2019-multifit}.
The cross-lingual music genre annotation would be closer to the XNLI task; hence we could fine-tune the pre-trained models on a parallel corpus of music genres translations or music genre annotations.
However, 
our research investigates language-specific semantic representations. 
Also, using translated music genres would not model their different perception across cultures while obtaining an exhaustive corpus of cross-lingual annotations is challenging.

Last, we explore \textbf{LASER}, a universal language-agnostic sentence embedding model \cite{Artetxe-Schwenk-2019}.
The model is based on a BiLSTM encoder trained on corpora in $93$ languages to learn multilingual fixed-length sentence embeddings.
As in other models, sentences are tokenized against a fixed vocabulary, obtained with BPE from the concatenated multilingual corpora.
\textbf{LASER} appears highly effective without requiring task-specific fine-tuning \cite{Artetxe-Schwenk-2019}.

\subsection{Retrofitting Music Genre Distributed Representations to Ontologies}
\label{sec:retro}
Retrofitting \cite{faruqui-etal-2015-retrofitting} is a method to refine vector space word representations by considering the relations between words as defined in semantic lexicons such as WordNet \cite{Miller_1995}.
The intuition is to modify the distributed embeddings to become closer to the representations of the concepts to which they are related.
Ever since the original work, many uses of retrofitting have been explored to semantically specialize word embeddings in relations such as synonyms or antonyms \cite{kiela-etal-2015-specializing,kim-etal-2016-adjusting}, in other languages than a source one \cite{ponti-etal-2019-cross} or in specific domains \cite{hangya-etal-2018-two}.

Enhanced extensions of retrofitting exist, but they require supervision \cite{lengerich-etal-2018-retrofitting}.
The original method \cite{faruqui-etal-2015-retrofitting} is unsupervised and can simply yet effectively leverage distributed embeddings and ontologies for improved representations.
Thus, we mainly rely on it, but we apply some changes as further described.
Let $\Omega=(C, E)$ be an ontology including the concepts $C$ and the semantic relations between these concepts $E \subseteq C \times C$.
The retrofitting goal is to learn new concept embeddings, $\textbf{Q} \in {\rm I\!R}^{n \times d}$ with $n=|C|$ and $d$ the embedding dimension.
The learning starts with initializing each $\textbf{q}_i \in {\rm I\!R}^d$, the new embedding for concept $i\in C$, to $\hat{\textbf{q}}_i$, the initial distributed embedding, and then iteratively updates $\textbf{q}_i$ until convergence as follows:
\begin{equation}
    \textbf{q}_i \leftarrow{} \frac{\sum_{j:(i,j) \in E}{(\beta_{ij}+ \beta_{ji})\textbf{q}_j} + \alpha_i \hat{\textbf{q}}_i}{\sum_{j:(i,j) \in E}{(\beta_{ij}+\beta_{ji})} + \alpha_i}
    \label{eq:update}
\end{equation}
$\alpha$ and $\beta$ are positive scalars weighting the importance of the initial, respectively, the related concept embeddings in computation.
The formula was reached through the optimization of the retrofitting objective using the Jacobi method \cite{saad2003iterative}.

\begin{table*}
\begin{small}
\centering
\begin{tabular}{|l|ll|llll|l|}
\hline
\textbf{Pair} & \textbf{GTrans} & \textbf{DBpSameAs} & \textbf{mBERT}$_{avg}$ & \textbf{FT}$_{sif}$ & \textbf{XLM}$_{Ctxt}$ & \textbf{LASER} & \textbf{Rfit}$_{u\Omega}$\textbf{FT}$_{sif}$\\
\hline
\textbf{en}-\textbf{nl} & 59.9 $\pm$ 0.3 & \underline{72.2 $\pm$ 0.2} & 86.2  $\pm$ 0.2 & \underline{86.5  $\pm$ 0.1} & 85.4  $\pm$ 0.3 & 80.9  $\pm$ 0.1 & \textbf{90.0  $\pm$ 0.1} \\
\textbf{en}-\textbf{fr} & 58.4 $\pm$ 0.1 & \underline{70.0  $\pm$ 0.2} & 85.2  $\pm$ 0.3 & \underline{87.4  $\pm$ 0.3} & 86.6  $\pm$ 0.2 & 82.0  $\pm$ 0.2 & \textbf{90.8  $\pm$ 0.2} \\
\textbf{en}-\textbf{es} & 56.9 $\pm$ 0.0 & \underline{65.4  $\pm$ 0.2} & 83.8  $\pm$ 0.1 & \underline{86.9  $\pm$ 0.2} & 85.4  $\pm$ 0.3 & 81.8  $\pm$ 0.1 & \textbf{89.9  $\pm$ 0.1} \\
\textbf{en}-\textbf{cs} & 60.6 $\pm$ 0.6 & \underline{78.4  $\pm$ 0.6} & \underline{88.2  $\pm$ 0.4} & \underline{88.6  $\pm$ 0.4} & \underline{89.0  $\pm$ 0.5} & 85.4  $\pm$ 0.3 & \textbf{90.4  $\pm$ 0.3} \\
\textbf{en}-\textbf{ja} & 60.9 $\pm$ 0.1 & \underline{70.4  $\pm$ 0.2} & 72.4  $\pm$ 0.2 & \underline{80.8  $\pm$ 0.3} & 74.2  $\pm$ 0.3 & 73.0  $\pm$ 0.1 & \textbf{86.7  $\pm$ 0.3} \\ \hline
\textbf{nl}-\textbf{en} & 53.5 $\pm$ 0.1 & \underline{56.7  $\pm$ 0.3} & 74.6  $\pm$ 0.3 & \underline{79.8  $\pm$ 0.4} & 74.2  $\pm$ 0.7 & 70.5  $\pm$ 0.1 & \textbf{84.3  $\pm$ 0.1} \\
\textbf{nl}-\textbf{fr} & 54.4 $\pm$ 0.2 & \underline{60.0  $\pm$ 0.4} & 76.1  $\pm$ 0.3 & \underline{79.3  $\pm$ 0.8} & 74.9  $\pm$ 1.0 & 72.6  $\pm$ 0.5 & \textbf{81.5  $\pm$ 0.7} \\
\textbf{nl}-\textbf{es} & 53.1 $\pm$ 0.2 & \underline{56.8  $\pm$ 0.2} & 75.0  $\pm$ 0.4 & \underline{77.7  $\pm$ 0.5} & 73.5  $\pm$ 0.3 & 71.0  $\pm$ 0.3 & \textbf{80.5  $\pm$ 0.4} \\
\textbf{nl}-\textbf{cs} & \underline{57.8 $\pm$ 0.1} & 50.0  $\pm$ 0.0 & \underline{80.8  $\pm$ 0.4} & \underline{80.6  $\pm$ 0.2} & 79.0  $\pm$ 0.4 & 76.8  $\pm$ 0.2 & \textbf{83.4  $\pm$ 0.5} \\
\textbf{nl}-\textbf{ja} & 57.5 $\pm$ 0.4 & \underline{62.7  $\pm$ 0.4} & 65.8  $\pm$ 2.5 & \underline{74.9  $\pm$ 1.0} & 66.0  $\pm$ 0.6 & 68.5  $\pm$ 0.2 & \textbf{80.0  $\pm$ 0.7} \\ \hline
\textbf{fr}-\textbf{nl} & 58.6 $\pm$ 0.1 & \underline{65.3  $\pm$ 0.4} & 79.4  $\pm$ 0.1 & \underline{81.9  $\pm$ 0.4} & 78.6  $\pm$ 0.1 & 74.3  $\pm$ 0.4 & \textbf{84.7  $\pm$ 0.3} \\
\textbf{fr}-\textbf{en} &  55.3 $\pm$ 0.0 & \underline{59.7  $\pm$ 0.2} & 77.2  $\pm$ 0.5 & \underline{83.0  $\pm$ 0.2} & 78.8  $\pm$ 0.5 & 74.2  $\pm$ 0.4 & \textbf{87.7  $\pm$ 0.1} \\
\textbf{fr}-\textbf{es} & 54.1 $\pm$ 0.1 & \underline{59.0  $\pm$ 0.1} & 77.5  $\pm$ 0.4 & \underline{81.8  $\pm$ 0.3} & 78.7  $\pm$ 0.5 & 75.4  $\pm$ 0.6 & \textbf{85.3  $\pm$ 0.2} \\
\textbf{fr}-\textbf{cs} & 59.1 $\pm$ 0.3 & \underline{70.0  $\pm$ 0.6} & 82.7  $\pm$ 0.6 & \underline{83.9  $\pm$ 0.4} & 83.1  $\pm$ 0.2 & 80.2  $\pm$ 0.5 & \textbf{87.2  $\pm$ 0.3} \\
\textbf{fr}-\textbf{ja} & 59.1 $\pm$ 0.2 & \underline{64.7  $\pm$ 0.5} & 61.5  $\pm$ 0.2 & \underline{77.9  $\pm$ 0.1} & 69.5  $\pm$ 0.3 & 70.5  $\pm$ 0.3 & \textbf{81.4  $\pm$ 0.3} \\ \hline
\textbf{es}-\textbf{nl} & 59.8 $\pm$ 0.3 & \underline{67.2  $\pm$ 0.2} & \underline{82.3  $\pm$ 0.3} & \underline{82.8  $\pm$ 0.9} & \underline{81.3  $\pm$ 0.8} & 76.7  $\pm$ 0.5 & \textbf{85.9  $\pm$ 0.6} \\
\textbf{es}-\textbf{fr} & 57.4 $\pm$ 0.2 & \underline{64.8  $\pm$ 0.3} & 81.0  $\pm$ 0.3 & \underline{85.0  $\pm$ 0.3} & 82.2  $\pm$ 0.5 & 78.1  $\pm$ 0.8 & \textbf{87.5  $\pm$ 0.3} \\
\textbf{es}-\textbf{en} & 57.0 $\pm$ 0.1 & \underline{61.7  $\pm$ 0.0} & 78.4  $\pm$ 0.3 & \underline{84.7  $\pm$ 0.2} & 79.5  $\pm$ 0.2 & 74.8  $\pm$ 0.4 & \textbf{88.8  $\pm$ 0.3} \\
\textbf{es}-\textbf{cs} & 60.3 $\pm$ 0.2 & \underline{72.2  $\pm$ 0.4} & \underline{85.5  $\pm$ 0.5} & \underline{85.6  $\pm$ 0.6} & \underline{85.9  $\pm$ 0.5} & 82.7  $\pm$ 0.5 & \textbf{88.0  $\pm$ 0.4} \\
\textbf{es}-\textbf{ja} & 60.9 $\pm$ 0.1 & \underline{67.0  $\pm$ 0.5} & 67.7  $\pm$ 0.1 & \underline{78.3  $\pm$ 0.6} & 72.8  $\pm$ 0.8 & 71.2  $\pm$ 0.4 & \textbf{83.1  $\pm$ 0.6} \\ \hline
\textbf{cs}-\textbf{nl} & \underline{57.6 $\pm$ 0.6} & 50.0  $\pm$ 0.0 & \underline{78.5  $\pm$ 0.7} & \underline{78.3 $\pm$ 0.9} & \underline{78.1  $\pm$ 0.4} & 73.1  $\pm$ 0.4 & \textbf{81.1  $\pm$ 1.2} \\
\textbf{cs}-\textbf{fr} & 54.2 $\pm$ 0.2 & \underline{60.0  $\pm$ 0.3} & 77.1  $\pm$ 1.0 & 78.5  $\pm$ 0.2 & \underline{79.9  $\pm$ 0.7} & 75.0  $\pm$ 0.9 & \textbf{81.4  $\pm$ 0.3} \\
\textbf{cs}-\textbf{es} & 53.7 $\pm$ 0.4 & \underline{56.9  $\pm$ 0.3} & 75.8  $\pm$ 0.3 & \underline{77.7  $\pm$ 0.8} & \underline{78.8  $\pm$ 0.4} & 73.7  $\pm$ 0.3 & \textbf{81.6  $\pm$ 0.9} \\
\textbf{cs}-\textbf{en} & 54.2 $\pm$ 0.2 & \underline{57.1  $\pm$ 0.1} & 74.2  $\pm$ 0.2 & \underline{78.9  $\pm$ 0.1} & 78.3  $\pm$ 0.4 & 73.4  $\pm$ 0.4 & \textbf{84.5  $\pm$ 0.4} \\
\textbf{cs}-\textbf{ja} & 58.6 $\pm$ 0.2 & \underline{64.0  $\pm$ 0.3} & 65.8  $\pm$ 1.4 & \underline{76.9  $\pm$ 0.1} & 72.2  $\pm$ 1.1 & 72.2  $\pm$ 1.1 & \textbf{80.5  $\pm$ 0.5} \\ \hline
\textbf{ja}-\textbf{nl} & 54.8 $\pm$ 0.4 & \underline{61.6  $\pm$ 1.1} & 62.1  $\pm$ 0.4 & \underline{72.8  $\pm$ 1.0} & 63.6  $\pm$ 0.9 & 68.3  $\pm$ 1.2 & \textbf{76.9  $\pm$ 0.3} \\
\textbf{ja}-\textbf{fr} & 53.3 $\pm$ 0.2 & \underline{58.4  $\pm$ 0.2} & 50.6  $\pm$ 0.2 & \underline{73.7  $\pm$ 0.6} & 58.4  $\pm$ 0.4 & 66.7  $\pm$ 0.3 & \textbf{77.8  $\pm$ 0.1} \\
\textbf{ja}-\textbf{es} & 52.7 $\pm$ 0.1 & \underline{55.9  $\pm$ 0.4} & 56.1  $\pm$ 0.4 & \underline{73.9  $\pm$ 0.4} & 60.0  $\pm$ 0.2 & 67.4  $\pm$ 0.4 & \textbf{78.8  $\pm$ 0.5} \\
\textbf{ja}-\textbf{cs} & 56.1 $\pm$ 0.5 & \underline{65.7  $\pm$ 0.7} & 64.7  $\pm$ 0.6 & \underline{77.5  $\pm$ 0.2} & 64.5  $\pm$ 0.5 & 73.3  $\pm$ 0.6 & \textbf{80.7  $\pm$ 0.4} \\
\textbf{ja}-\textbf{en} & 52.5 $\pm$ 0.1 & \underline{55.8  $\pm$ 0.1} & 49.0  $\pm$ 0.1 & \underline{75.6  $\pm$ 0.3} & 56.8  $\pm$ 1.0 & 64.2  $\pm$ 0.4  & \textbf{81.6  $\pm$ 0.8} \\
\hline
\end{tabular}
\caption{\label{tab:results}
Macro-AUC scores (in \%, best overall in bold, best locally underlined).
The first part corresponds to the translation baselines;
the second to the best distributed representations;
the last to the retrofitted $\textbf{FT}_{sif}$ vectors.
}
\end{small}
\end{table*}

Equation \ref{eq:update} is a corrected version of the original work, as for a concept $i$, not only $\beta_{ij}$ appears in it, but also $\beta_{ji}$.
That is to say that when computing the partial derivative of the retrofitting objective concerning $i$, two non-zero terms are corresponding to the related concept $j$: when $i$ is the source and $j$ is the target and vice-versa \cite{bengio2006label,saha2016dis}.
The further modifications that we make regard the parameters $\alpha$ and $\beta$.
For each $i\in C$, \citet{faruqui-etal-2015-retrofitting} fix $\alpha_i$ to $1$, and $\beta_{ij}$ to $\frac{1}{degree(i)}$ for $(i,j) \in E$ or $0$ otherwise; $degree(i)$ is the number of related concepts $i$ has in $\Omega$.

While many embedding models can handle unknown words nowadays, concepts may still have unknown initial distributed vectors, depending on the model's choice.
For this case, expanded retrofitting \cite{speer2016ensemble} has been proposed, considering $\alpha_i=0$, for each concept $i$ with unknown initial distributed vector, and $\alpha_i=1$ for the rest.
Thus, $\textbf{q}_i$ is initialized to $\textbf{0}_d$ and updated by averaging the embeddings of its related concepts at each iteration.
Let us notice that, through retrofitting, representations are not only modified but also learned from scratch for some concepts.

We adopt the same approach to initialize $\alpha$.
Moreover, we also adjust the parameters $\beta$ to weight the importance of each related concept embedding depending on the relation semantics in our music genre ontology \cite{Epure2020}.
Specifically, we distinguish between equivalence and relatedness as follows:
\[
\overline{\beta}_{ij} = \left\{
        \begin{array}{ll}
        1 & : (i,j) \in E_\epsilon \subset E\\
        \beta_{ij} & : (i,j) \in E - E_\epsilon\\ 
        0, & : (i,j) \not\in E
        \end{array}
    \right.
\]
where $E_\epsilon$ contains the equivalence relation types (\textit{wikiPageRedirects}, \textit{sameAs}); 
$E - E_\epsilon$ contains the relatedness relation types (\textit{stylisticOrigin}, \textit{musicSubgenre}, \textit{derivative}, \textit{musicFusionGenre}).
We label this modified version of retrofitting as \textbf{Rfit}.

Finally, we want to highlight a crucial aspect of retrofitting.
Previous works \cite{speer2016ensemble,hayes-2019-just,Fang2019} claim that, while the retrofitting updating procedure converges, the results depend on the order in which the updates are made. 
We prove in Appendix \ref{app:convex} that the retrofitting objective is strictly convex when at least one initial concept vector is known in each connected component.
Hence, with this condition satisfied, retrofitting converges to the same solution always and independently of the updates' order.

\section{Experiments}
\label{sec:results}
Cross-lingual music genre annotation, as formalized in Section \ref{sec:crossannotation}, is a typical multi-label prediction task.
For evaluation, we use the Area Under the receiver operating characteristic Curve (AUC, \citealp{BRADLEY19971145}), macro-averaged.
We report the mean and standard deviations of the macro AUC scores using $3$-fold cross-validation.
For each language, we apply an iterative split \cite{Sechidis:2011} of the test corpus that balances the number of samples and the tag distributions across the folds.
We pre-process the music genres by either replacing special characters with space (\textit{\_-/,}) or removing them (\textit{()':.!\$}).
For Japanese, we introduce spaces between tokens obtained with Mecab\footnote{\url{https://taku910.github.io/mecab/}}.
Embeddings are then computed from pre-processed tags.

We test two translation baselines, one based on Google Translate\footnote{\url{https://translate.google.com}} (\textbf{GTrans}) and one on the DBpedia \textit{SameAs} relation (\textbf{DBpSameAs}).
In this case, a source music genre is mapped on a single or no target music genre, its embedding being in the form $\textbf{\{0,1\}}^{|T|}$.
For \textbf{XLM}$_{Ctxt}$, we compute the sentence embedding by averaging the token embeddings obtained with mean pooling across all layers.
For \textbf{mBERT}$_{Ctxt}$, we apply the same strategy, but by max pooling the token embeddings instead.
We chose these representations as they showed the best performance experimentally compared to the other strategies described in Section \ref{sec:embs}.

When retrofitting language-specific music genre embeddings, we use the corresponding monolingual ontology (\textbf{Rfit}$_{u\Omega}$).
When we learn multilingual embeddings from scratch with retrofitting, by knowing only music genre embeddings in one language (\textit{la}), we use the partially aligned DBpedia ontologies which contain the \textit{SameAs} relations (\textbf{Rfit}$_{a\Omega}^{la}$).
For this case, we also propose a baseline representing a source concept embedding as a vector of geodesic distances in the partially aligned ontologies to each target concept  (\textbf{DBp$_{a\Omega}$NNDist}).

\paragraph{Results.}
Table \ref{tab:results} shows the cross-lingual annotation results.
The standard translation, \textbf{GTrans}, leads to the lowest results being over-performed by a knowledge-based translation, more adapted to this domain (\textbf{DBpSameAs}).
Also, these results show that translation methods fail to capture the dissimilar cross-cultural music genre perception.

The second part of Table \ref{tab:results} contains only the best\footnote{The complete results are presented in Appendix \ref{app:results}.} music genre embeddings computed with each word/sentence pre-trained model or method.
When averaging static multilingual word embeddings, those from \textbf{mBERT} often yield the most relevant cross-lingual annotations, while when applying the \textit{sif} averaging, the aligned \textbf{FT} word vectors are the best choice.
Between the two contextual word embedding models, \textbf{XLM}$_{Ctxt}$ significantly outperforms \textbf{mBERT}$_{Ctxt}$, thus we report only the former.

We can notice that all distributed representations of music genres can model quite well the varying music genre annotation across languages.
\textbf{FT}$_{sif}$ results in the most relevant cross-lingual annotations consistently for $5$ out of $6$ languages as a source.
For \textbf{cs} though, the embeddings from \textbf{XLM}$_{Ctxt}$ are sometimes slightly better.
\textbf{LASER} under-performs for most languages but \textbf{ja}, for which the vectors obtained with \textbf{mBERT}$_{avg}$ are less suitable. 

The last column of Table \ref{tab:results} shows the results of cross-lingual annotation when using the \textbf{FT}$_{sif}$ vectors retrofitted to monolingual music genre ontologies.
The domain adaptation of concept embeddings, inferred with general-language pre-trained models, significantly improves music genre annotation modeling across all pairs of languages.

\begin{table}
\begin{small}
\centering
\begin{tabular}{|l|l|l|}
\hline
\textbf{Pair} & \textbf{DBp$_{a\Omega}$NNDist} & \textbf{Rfit$_{a\Omega}^{en}$}\textbf{FT}$_{sif}$  \\
\hline
\textbf{en}-\textbf{nl} & 83.5  $\pm$ 0.1 & \textbf{90.7  $\pm$ 0.0} \\
\textbf{en}-\textbf{fr} & 82.7  $\pm$ 0.3 & \textbf{91.7  $\pm$ 0.1} \\
\textbf{en}-\textbf{es} & 81.1  $\pm$ 0.3 & \textbf{91.4  $\pm$ 0.2} \\
\textbf{en}-\textbf{cs} & 86.6  $\pm$ 0.3 & \textbf{91.6  $\pm$ 0.4} \\
\textbf{en}-\textbf{ja} & 81.3  $\pm$ 0.1 & \textbf{89.4  $\pm$ 0.2} \\
\hline
\textbf{Pair} & \textbf{DBp$_{a\Omega}$NNDist} & \textbf{Rfit$_{a\Omega}^{ja}$}\textbf{FT}$_{sif}$  \\
\hline
\textbf{ja}-\textbf{nl} & 68.2  $\pm$ 0.7 & 75.7  $\pm$ 0.3 \\
\textbf{ja}-\textbf{fr} & 71.9  $\pm$ 0.1 & 76.7  $\pm$ 0.5 \\
\textbf{ja}-\textbf{es} & 68.9  $\pm$ 0.4 & 76.1  $\pm$ 0.5 \\
\textbf{ja}-\textbf{cs} & 77.9  $\pm$ 0.8 & \textbf{82.2  $\pm$ 0.8} \\
\textbf{ja}-\textbf{en} & 70.4  $\pm$ 0.4 & \textbf{82.3  $\pm$ 0.5} \\
\hline
\end{tabular}
\caption{\label{tab:retrofit_la_only}
Macro-AUC scores (in \%; those larger than \textbf{Rfit$_{u\Omega}$}\textbf{FT}$_{sif}$ in Table \ref{tab:results} in bold) with vectors learned by retrofitting to aligned monolingual ontologies.
}
\end{small}
\end{table}

Table \ref{tab:retrofit_la_only} shows the results when using retrofitting to learn music genre embeddings from scratch.
Here, distributed vectors are known for one language  (\textbf{en}, respectively \textbf{ja}\footnote{The results for the other languages are in Appendix \ref{app:results}.}) and the monolingual ontologies are partially aligned.
Even though not necessarily all music genres are linked to their equivalents in the other language, the concept representations learned in this way are more relevant for cross-lingual annotation, for all pairs involving \textbf{en} as the source and for \textbf{ja}-\textbf{cs} and \textbf{ja}-\textbf{en}.
In fact, the baseline (\textbf{DBp$_{a\Omega}$NNDist}) reveals that the aligned ontologies stand-alone can model the cross-lingual annotation quite well, in particular for \textbf{en}.

\paragraph{Discussion.}
\begin{CJK}{UTF8}{ipxm}
The results show that using translation to produce cross-lingual annotations is limited as it does not consider the culturally divergent perception of music genres.
Instead, monolingual semantic representations can model this phenomenon rather well.
For instance, from \emph{Milton Cardona}'s music genres in \textbf{es}, \emph{salsa} and \emph{jazz}, it correctly predicts the Japanese equivalent of \emph{fusion} (フュージョン) in \textbf{ja}.
Yet, while a thorough qualitative analysis requires more work, preliminary exploration suggests that larger gaps in perception might still be inadequately modeled.
For instance, for \emph{Santana}'s album \emph{Welcome} tagged with \emph{jazz} in \textbf{es}, it does not predict \emph{pop} in \textbf{fr}.
\end{CJK}

When comparing the distributed embeddings, a simple method that relies on a weighted average of multilingual aligned word vectors significantly outperforms the others.
Although rarely used before, we question if we can notice such high performance with other multilingual data-sets.
The cross-lingual annotations are further improved by retrofitting the distributed embeddings to monolingual ontologies.
Interestingly, the vector alignment does not appear degraded by retrofitting to disjoint graphs.
Or, the negative impact is limited and exceeded by introducing domain knowledge in representations.
Further, as shown in Table \ref{tab:retrofit_la_only}, joining semantic representations in this way proves very suitable to learn music genre vectors from scratch.

Regarding the scores per language, we obtained the lowest ones for \textbf{ja} as the source.
We could explain this by either a more challenging test corpus or still incompatible embeddings in \textbf{ja}, possibly because of the quality of the individual embedding models for this language and the completeness of the Japanese music genre ontology. 
Also, we did not notice any particular improvement for pairs of languages from the same language family, \textit{e.g.} \textbf{fr} and \textbf{es}.
However, we would need a sufficiently sizeable parallel corpus exhaustively annotated in all languages to reliably compare the performance for pairs of languages from the same language family or different ones.

Finally, by closely analysing the results in Table \ref{tab:results}, we noticed that given two languages $L_1$ and $L_2$, with more music genre embeddings in $L_1$ than in $L_2$ (from both ontology and corpus), the results of mapping annotations from $L_1$ to $L_1$ seems always better than the results from $L_2$ to $L_1$.
This observation explains two trends in Table \ref{tab:results}.
First, the scores achieved for \textbf{en} or \textbf{es} as the source, the languages with the largest number of music genres, are the best.
Second, the results for the same pair of languages could vary a lot, depending on the role each language plays, source, or target.

One possible explanation is that the prediction from languages with fewer music genre tags such as $L_2$ towards languages with more music genre tags such as $L_1$ is more challenging because the target language contains more specific or rare annotations.
For instance, when checking the results per tag from \textbf{cs} to \textbf{en} we observed that among the tags with the lowest scores, we found \textit{moombahton}, \textit{zeuhl}, or \textit{candombe}.
However, other common music genres, such as \textit{latin music} or \textit{hard rock}, were also poorly predicted, showing that other causes exist too.
Is the unbalanced number of music genres used in annotations a cultural consequence? 
Related work \cite{Ferwerda2016InvestigatingTR} seems to support this hypothesis.
Then could we design a better mapping function that leverages the unbalanced numbers of music genres in cross-cultural annotations? 
We will dedicate a thorough investigation of these questions as future work.

\section{Conclusion}
\label{sec:conclusion}
We have presented an extensive investigation on cross-lingual modeling of music genre annotation, focused on six languages, and two common approaches to semantically represent concepts: ontologies and distributed embeddings\footnote{\url{https://github.com/deezer/CrossCulturalMusicGenrePerception}}.

Our work provides a methodological framework to study the annotation behavior across language-bound cultures in other domains too.
Hence, the effectiveness of language-specific concept representations to model the culturally diverse perception could be further probed.
Then, we combined the semantic representations only with retrofitting.
However, inspired by paraphrastic sentence embedding learning, one can also consider the music genre relations as paraphrasing forms with different strengths \cite{Wieting2016}.
Finally, the models to generate cross-lingual annotations should be thoroughly evaluated in downstream music retrieval and recommendation tasks.

\bibliography{anthology,emnlp2020}
\bibliographystyle{acl_natbib}

\newpage
\appendix
\section{Strict Convexity of Retrofitting}
\label{app:convex}
%Further, we use the same notation as in the original retrofitting work \cite{faruqui-etal-2015-retrofitting}.

\paragraph{Theorem.} Let $V$ be a finite vocabulary with $|V| = n$.
Let $\Omega = (V,E)$ be an ontology represented as a directed graph which encodes semantic relationships between vocabulary words.
Further, let $\hat{V} \subseteq V$ be the subset of words which have non-zero initial distributed representations, $\hat{\textbf{q}}_i$. 
The goal of retrofitting is to learn the matrix $\textbf{Q} \in {\rm I\!R}^d$, stacking up the new embeddings $\textbf{q}_i \in {\rm I\!R}^d$ for each $i \in V$.
The objective function to be minimized is:
\[
\begin{array}{l}
\Phi(\textbf{Q}) = \sum_{i \in \hat{V}}\alpha_i ||\textbf{q}_i - \hat{\textbf{q}}_i||^2_2 \\
+ \sum_{i =1}^{n} \sum_{(i, j)\in E}{\beta_{ij}||\textbf{q}_i - \textbf{q}_j||^2_2 },
\end{array}
\]
where the $\alpha_i$ and $\beta_{ij}$ are positive scalars. 
Assuming that each connected component of $\Omega$ includes at least one word from $\hat{V}$, the objective function $\Phi$ is strictly convex w.r.t. $\textbf{Q}$.

\paragraph{Proof.} 
First of all, let $\hat{\textbf{Q}}$ denote the $n \times d$ matrix whose $i$-th row corresponds to $\hat{\textbf{q}}_i$ if $i \in \hat{V}$, and to the $d$-dimensional null vector $\textbf{0}_d$ otherwise. 
Let $\textbf{A}$ denote the $n \times n$ diagonal matrix verifying $\textbf{A}_{ii} = \alpha_i$ if $i \in \hat{V}$ and $\textbf{A}_{ii} = 0$ otherwise. 
Let $\textbf{B}$ denote the $n\times n$ symmetric matrix such as, for all $i, j \in \{1,...,n\}$ with $i \neq j$, $\textbf{B}_{ij} = \textbf{B}_{ji} = -\frac{1}{2} (\beta_{ij} + \beta_{ji})$ and $\textbf{B}_{ii} = \sum_{j=1,j\neq i}^n |\textbf{B}_{ij}|$. 
With these notations, and with $Tr(\cdot)$ the trace operator for square matrices, we have:
\[
\begin{array}{l}
\sum_{i \in \hat{V}}\alpha_i ||\textbf{q}_i - \hat{\textbf{q}}_i||^2_2 \\
= Tr\Big( (\textbf{Q} - \hat{\textbf{Q}})^T \textbf{A} (\textbf{Q} - \hat{\textbf{Q}}) \Big) \\
= Tr\Big(\textbf{Q}^T \textbf{A} \textbf{Q} - \hat{\textbf{Q}}^T \textbf{A} \textbf{Q} - \textbf{Q}^T \textbf{A} \hat{\textbf{Q}} + \hat{\textbf{Q}}^T \textbf{A} \hat{\textbf{Q}}\Big).
\end{array}
\]
Also:
$$\sum_{i =1}^{n} \sum_{(i, j)\in E}{\beta_{ij}||\textbf{q}_i - \textbf{q}_j||^2_2 } = Tr\Big( \textbf{Q}^T \textbf{B} \textbf{Q} \Big).$$
Therefore, as the trace is a linear mapping, we have:
\[
\begin{array}{l}
\Phi(\textbf{Q}) = Tr\Big( \textbf{Q}^T (\textbf{A} + \textbf{B})\textbf{Q}\Big) + \\ Tr\Big(\hat{\textbf{Q}}^T \textbf{A} \hat{\textbf{Q}} - \hat{\textbf{Q}}^T \textbf{A} \textbf{Q} - \textbf{Q}^T \textbf{A} \hat{\textbf{Q}}\Big).
\end{array}
\]
Then, we note that $\textbf{A} + \textbf{B}$ is a weakly diagonally dominant matrix (WDD) as, by construction, $\forall i \in \{1,...,n\}, |(\textbf{A}+\textbf{B})_{ii}| \geq \sum_{j \neq i} |(\textbf{A}+\textbf{B})_{ij}|$. 
Also, for all $i \in \hat{V}$, the inequality is strict, as $|(\textbf{A}+\textbf{B})_{ii}| = \alpha_i + \sum_{j \neq i} |\textbf{B}_{ij}| > \sum_{j \neq i} |(\textbf{A}+\textbf{B})_{ij}| = \sum_{j \neq i} |\textbf{B}_{ij}|$, which means that, for all $i \in \hat{V}$, row $i$ of $\textbf{A} + \textbf{B}$ is strictly diagonally dominant (SSD). 
Assuming that each connected component of graph $G$ includes at least one node from $\hat{V}$, we conclude that $\textbf{A} +\textbf{B}$ is a weakly chained diagonally dominant matrix \cite{azimzadeh2016weakly}, \emph{i.e.} that:
\begin{itemize}
    \item $\textbf{A}+\textbf{B}$ is WDD;
    \item for each $i \in V$ such that row $i$ is not SSD, there exists a \textit{walk} in the graph whose adjacency matrix is $\textbf{A}+\textbf{B}$ (two nodes $i$ and $j$ are connected if $(\textbf{A}+\textbf{B})_{ij} = (\textbf{A}+\textbf{B})_{ji} \neq 0$), starting from $i$ and ending at a node associated to a SSD row.
\end{itemize}
Such matrices are nonsingular \cite{azimzadeh2016weakly}, which implies that $\textbf{Q} \rightarrow \textbf{Q}^T (\textbf{A} + \textbf{B}) \textbf{Q}$ is a positive-definite quadratic form. As $\textbf{A} + \textbf{B}$ is a symmetric positive-definite matrix, there exists a matrix $\textbf{M}$ such that $\textbf{A} + \textbf{B} = \textbf{M}^T \textbf{M}$. Therefore, denoting $||\cdot||_F^2$ the squared Frobenius matrix norm:
\[
\begin{array}{l}
Tr\Big( \textbf{Q}^T (\textbf{A} + \textbf{B}) \textbf{Q} \Big) \\ 
= Tr\Big( \textbf{Q}^T  \textbf{M}^T \textbf{M} \textbf{Q} \Big)
= || \textbf{Q} \textbf{M} ||_F^2
\end{array}
\]
which is strictly convex w.r.t. $\textbf{Q}$ due to the strict convexity of the squared Frobenius norm (see \emph{e.g.} 3.1 in \citet{dattorro2005convex}). 
Since the sum of strictly convex functions of $\textbf{Q}$ (first trace in $\Phi(\textbf{Q})$) and linear functions of $\textbf{Q}$ (second trace in $\Phi(\textbf{Q})$) is still strictly convex w.r.t. $\textbf{Q}$, we conclude that the objective function $\Phi$ is strictly convex w.r.t. $\textbf{Q}$.

\paragraph{Corollary.}
\textit{The retrofitting update procedure is insensitive to the order in which nodes are updated.}

The aforementioned updating procedure for $\textbf{Q}$ \cite{faruqui-etal-2015-retrofitting} is derived from Jacobi iteration procedure \cite{saad2003iterative, bengio2006label} and converges for any initialization.
Such a convergence result is discussed in \citet{bengio2006label}. 
It can also be directly verified in our specific setting by checking that each irreducible element of $\textbf{A}+\textbf{B}$, \emph{i.e.} each connected component of the underlying graph constructed from this matrix, is irreducibly diagonally
dominant (see 4.2.3 in \citet{saad2003iterative}) and then by applying Theorem 4.9 from \citet{saad2003iterative} on each of these components.
Besides, due to its strict convexity w.r.t. $\textbf{Q}$, the objective function $\Phi$ admits a unique global minimum. 
Consequently, the retrofitting update procedure will converge to the same embedding matrix regardless of the order in which nodes are updated.

\section{Extended Results}
The following Tables 5 and 6 provide more complete results from our experiments.
\label{app:results}

\begin{table*}[!htbp]
\begin{small}
\centering
\begin{tabular}{|l|lll|lll|ll|}
\hline
\textbf{Pair} & \textbf{FT}$_{avg}$ & \textbf{XLM}$_{avg}$ & \textbf{mBERT}$_{avg}$ & \textbf{FT}$_{sif}$ & \textbf{XLM}$_{sif}$ & \textbf{mBERT}$_{sif}$ & \textbf{XLM}$_{ctxt}$ & \textbf{mBERT}$_{ctxt}$\\
\hline
\textbf{en}-\textbf{nl} & 75.2 $\pm$ 0.2 & 83.7 $\pm$ 0.1 & \underline{86.2 $\pm$ 0.2} & 86.5 $\pm$ 0.1 & 86.6 $\pm$ 0.2 & \underline{88.3 $\pm$ 0.2} & \underline{85.4 $\pm$ 0.3} & \underline{85.1 $\pm$ 0.2} \\
\textbf{en}-\textbf{fr} & 78.6 $\pm$ 0.2 & 84.4 $\pm$ 0.2 & \underline{85.2 $\pm$ 0.3} & \underline{87.4 $\pm$ 0.3} & \underline{87.7 $\pm$ 0.3} & \underline{87.6 $\pm$ 0.2} & \underline{86.6 $\pm$ 0.2} & 84.9 $\pm$ 0.3 \\
\textbf{en}-\textbf{es} & 77.5 $\pm$ 0.1 & 82.5 $\pm$ 0.1 & \underline{83.8 $\pm$ 0.1} & \underline{86.9 $\pm$ 0.2} & \underline{86.7 $\pm$ 0.2} & 86.2 $\pm$ 0.2 & \underline{85.4 $\pm$ 0.3} & 84.7 $\pm$ 0.2 \\
\textbf{en}-\textbf{cs} & 74.5 $\pm$ 0.6 & 87.2 $\pm$ 0.6 & \underline{88.2 $\pm$ 0.4} & 88.6 $\pm$ 0.4 & \underline{90.6 $\pm$ 0.5} & \underline{90.6 $\pm$ 0.3} & \underline{89.0 $\pm$ 0.5} & 87.2 $\pm$ 0.1 \\
\textbf{en}-\textbf{ja} & 69.6 $\pm$ 0.5 & 70.9 $\pm$ 0.1 & \underline{72.4 $\pm$ 0.2} & \underline{80.8 $\pm$ 0.3} & 76.1 $\pm$ 0.3 & 70.5 $\pm$ 0.3 & \underline{74.2 $\pm$ 0.3} & 68.9 $\pm$ 0.2 \\
\textbf{nl}-\textbf{en} & 73.8 $\pm$ 0.5 & 71.9 $\pm$ 0.4 & \underline{74.6 $\pm$ 0.3} & \underline{79.8 $\pm$ 0.4} & 77.7 $\pm$ 0.4 & 78.2 $\pm$ 0.3 & \underline{74.2 $\pm$ 0.7} & \underline{73.5 $\pm$ 0.5} \\
\textbf{nl}-\textbf{fr} & 63.9 $\pm$ 0.5 & 72.7 $\pm$ 0.9 & \underline{76.1 $\pm$ 0.3} & \underline{79.3 $\pm$ 0.8} & 78.4 $\pm$ 1.0 & \underline{79.5 $\pm$ 0.3} & \underline{74.9 $\pm$ 1.0} & \underline{75.1 $\pm$ 0.7} \\
\textbf{nl}-\textbf{es} & 63.7 $\pm$ 0.4 & 71.6 $\pm$ 0.6 & \underline{75.0 $\pm$ 0.4} & \underline{77.7 $\pm$ 0.5} & 76.8 $\pm$ 0.3 & \underline{77.6 $\pm$ 0.6} & 73.5 $\pm$ 0.3 & \underline{75.0 $\pm$ 0.3} \\
\textbf{nl}-\textbf{cs} & 65.1 $\pm$ 0.3 & 77.4 $\pm$ 0.4 & \underline{80.8 $\pm$ 0.4} & 80.6 $\pm$ 0.2 & 82.0 $\pm$ 0.4 & \underline{83.2 $\pm$ 0.4} & 79.0 $\pm$ 0.4 & \underline{79.6 $\pm$ 0.2} \\
\textbf{nl}-\textbf{ja} & 64.8 $\pm$ 0.2 & \underline{65.2 $\pm$ 0.8} & \underline{65.8 $\pm$ 2.5} & \underline{74.9 $\pm$ 1.0} & 70.1 $\pm$ 0.4 & 67.6 $\pm$ 0.2 & \underline{66.0 $\pm$ 0.6} & 65.0 $\pm$ 0.2 \\
\textbf{fr}-\textbf{nl} & 67.7 $\pm$ 1.0 & 77.5 $\pm$ 0.4 & \underline{79.4 $\pm$ 0.1} & 81.9 $\pm$ 0.4 & 81.6 $\pm$ 0.3 & \underline{82.0 $\pm$ 0.4} & \underline{78.6 $\pm$ 0.1} & 77.7 $\pm$ 0.2 \\
\textbf{fr}-\textbf{en} & 76.2 $\pm$ 0.2 & 74.7 $\pm$ 0.6 & \underline{77.2 $\pm$ 0.5} & \underline{83.0 $\pm$ 0.2} & 81.5 $\pm$ 0.3 & 80.8 $\pm$ 0.1 & \underline{78.8 $\pm$ 0.5} & 77.2 $\pm$ 0.5 \\
\textbf{fr}-\textbf{es} & 71.0 $\pm$ 0.2 & 75.6 $\pm$ 0.3 & \underline{77.5 $\pm$ 0.4} &  \underline{81.8 $\pm$ 0.3} & 81.0 $\pm$ 0.5 & 80.0 $\pm$ 0.4 & \underline{78.7 $\pm$ 0.5} & \underline{78.2 $\pm$ 0.3} \\
\textbf{fr}-\textbf{cs} & 70.4 $\pm$ 0.8 & 80.1 $\pm$ 0.4 & \underline{82.7 $\pm$ 0.6} & 83.9 $\pm$ 0.4 & \underline{85.3 $\pm$ 0.3} & \underline{85.5 $\pm$ 0.5} & \underline{83.1 $\pm$ 0.2} & 82.0 $\pm$ 0.4 \\
\textbf{fr}-\textbf{ja} & \underline{71.1 $\pm$ 0.3} & 67.4 $\pm$ 0.5 & 61.5 $\pm$ 0.2 & \underline{77.9 $\pm$ 0.1} & 73.1 $\pm$ 0.1 & 66.9 $\pm$ 0.2 & \underline{69.5 $\pm$ 0.3} & 64.8 $\pm$ 0.5 \\
\textbf{es}-\textbf{nl} & 68.5 $\pm$ 0.5 & 80.3 $\pm$ 0.6 & \underline{82.3 $\pm$ 0.3} & 82.8 $\pm$ 0.9 & 83.8 $\pm$ 0.7 & \underline{84.8 $\pm$ 0.2} & \underline{81.3 $\pm$ 0.8} & \underline{80.7 $\pm$ 0.4} \\
\textbf{es}-\textbf{fr} & 70.8 $\pm$ 0.4 & 79.9 $\pm$ 0.5 & \underline{81.0 $\pm$ 0.3} & \underline{85.0 $\pm$ 0.3} & \underline{84.6 $\pm$ 0.4} & \underline{84.5 $\pm$ 0.4} & \underline{82.2 $\pm$ 0.5} & 80.3 $\pm$ 0.4 \\
\textbf{es}-\textbf{en} & 75.3 $\pm$ 0.1 & 76.4 $\pm$ 0.2 & \underline{78.4 $\pm$ 0.3} & \underline{84.7 $\pm$ 0.2} & 83.2 $\pm$ 0.4 & 82.9 $\pm$ 0.3 & \underline{79.5 $\pm$ 0.2} & 78.1 $\pm$ 0.4 \\
\textbf{es}-\textbf{cs} & 68.9 $\pm$ 0.8 & 83.2 $\pm$ 0.7 & \underline{85.5 $\pm$ 0.5} & 85.6 $\pm$ 0.6 & \underline{88.2 $\pm$ 0.7} & \underline{88.6 $\pm$ 0.6} & \underline{85.9 $\pm$ 0.5} & 84.2 $\pm$ 0.4 \\
\textbf{es}-\textbf{ja} & 65.2 $\pm$ 0.4 & \underline{70.4 $\pm$ 0.2} & 67.7 $\pm$ 0.1 & \underline{78.3 $\pm$ 0.6} & 74.4 $\pm$ 0.2 & 68.6 $\pm$ 0.5 & \underline{72.8 $\pm$ 0.8} & 65.1 $\pm$ 0.6 \\
\textbf{cs}-\textbf{nl} & 68.5 $\pm$ 1.0 & 75.7 $\pm$ 0.9 & \underline{78.5 $\pm$ 0.7} & 78.3 $\pm$ 0.9 & \underline{80.7 $\pm$ 0.9} & \underline{80.5 $\pm$ 1.0} & \underline{78.1 $\pm$ 0.4} & 76.8 $\pm$ 0.3 \\
\textbf{cs}-\textbf{fr} & 64.5 $\pm$ 0.9 & 74.7 $\pm$ 0.9 & \underline{77.1 $\pm$ 1.0} & 78.5 $\pm$ 0.2 & \underline{80.7 $\pm$ 0.3} & \underline{80.7 $\pm$ 0.9} & \underline{79.9 $\pm$ 0.7} & 76.2 $\pm$ 1.2 \\
\textbf{cs}-\textbf{es} & 65.3 $\pm$ 0.9 & 73.8 $\pm$ 0.7 & \underline{75.8 $\pm$ 0.3} & 77.7 $\pm$ 0.8 & \underline{79.8 $\pm$ 0.2} & 78.7 $\pm$ 0.3 & \underline{78.8 $\pm$ 0.4} & 75.9 $\pm$ 0.8 \\
\textbf{cs}-\textbf{en} & 70.3 $\pm$ 0.5  & 70.9 $\pm$ 0.0 & \underline{74.2 $\pm$ 0.2} & \underline{78.9 $\pm$ 0.1} & \underline{78.9 $\pm$ 0.5} & \underline{79.2 $\pm$ 0.5} & \underline{78.3 $\pm$ 0.4} & 74.0 $\pm$ 0.3 \\
\textbf{cs}-\textbf{ja} &  67.1 $\pm$ 1.1 & \underline{70.8 $\pm$ 0.2} & 65.8 $\pm$ 1.4 & \underline{76.9 $\pm$ 0.1} & 72.7 $\pm$ 0.3 & 67.3 $\pm$ 0.7 & \underline{72.2 $\pm$ 1.1} & 68.3 $\pm$ 0.1 \\
\textbf{ja}-\textbf{nl} & \underline{62.0 $\pm$ 0.5} & 59.5 $\pm$ 1.0 & \underline{62.1 $\pm$ 0.4} & \underline{72.8 $\pm$ 1.0} & 68.0 $\pm$ 0.6 & 67.1 $\pm$ 0.2 & \underline{63.6 $\pm$ 0.9} & 52.4 $\pm$ 0.3 \\
\textbf{ja}-\textbf{fr} & \underline{66.0 $\pm$ 1.1} & 47.6 $\pm$ 0.2 & 50.6 $\pm$ 0.2 & \underline{73.7 $\pm$ 0.6} & 69.4 $\pm$ 0.7 & 65.6 $\pm$ 0.3 & \underline{58.4 $\pm$ 0.4} & 43.9 $\pm$ 0.7 \\
\textbf{ja}-\textbf{es} & \underline{63.2 $\pm$ 0.3} & 53.6 $\pm$ 0.3 & 56.1 $\pm$ 0.4 & \underline{73.9 $\pm$ 0.4} & 67.5 $\pm$ 1.0 & 63.6 $\pm$ 0.6 & \underline{60.0 $\pm$ 0.2} & 48.7 $\pm$ 0.6 \\
\textbf{ja}-\textbf{cs} & 61.7 $\pm$ 1.3 & \underline{64.8 $\pm$ 0.8} & \underline{64.7 $\pm$ 0.6} & \underline{77.5 $\pm$ 0.2} & 73.3 $\pm$ 0.6 & 69.2 $\pm$ 0.4 & \underline{64.5 $\pm$ 0.5} & 56.8 $\pm$ 0.9 \\
\textbf{ja}-\textbf{en} & \underline{72.1 $\pm$ 1.0} & 46.3 $\pm$ 0.4 & 49.0 $\pm$ 0.1 & \underline{75.6 $\pm$ 0.3} & 66.8 $\pm$ 0.6 & 64.1 $\pm$ 0.7 & \underline{56.8 $\pm$ 1.0} & 44.0 $\pm$ 0.2 \\
\hline
\end{tabular}
\caption{\label{results2}
Macro-AUC scores (in \%, best locally underlined).
The first two parts correspond to averaging or applying \textit{sif} averaging to static multilingual word embeddings;
the third part corresponds to the contextual sentence embeddings.
}
\end{small}
\end{table*}

\begin{table}
\begin{small}
\centering
\begin{tabular}{|l|l|lll|}
\hline
\textbf{Pair} & \textbf{DBp$_{a\Omega}$NNDist} & \textbf{Rfit}$_{u\Omega}$\textbf{FT}$_{sif}$ & \textbf{Rfit$_{a\Omega}^{source}$}\textbf{FT}$_{sif}$ & \textbf{Rfit$_{a\Omega}^{target}$}\textbf{FT}$_{sif}$ \\
\hline
\textbf{en}-\textbf{nl} & 83.5 $\pm$ 0.1 & 90.0 $\pm$ 0.1 & \textbf{90.7 $\pm$ 0.0} & \textbf{91.8  $\pm$ 0.2} \\
\textbf{en}-\textbf{fr} & 82.7 $\pm$ 0.3 & 90.8 $\pm$ 0.2 & \textbf{91.7 $\pm$ 0.1} & \textbf{91.8  $\pm$ 0.2} \\
\textbf{en}-\textbf{es} & 81.1 $\pm$ 0.3 & 89.9 $\pm$ 0.1 & \textbf{91.4 $\pm$ 0.2} & \textbf{90.9  $\pm$ 0.2} \\
\textbf{en}-\textbf{cs} & 86.6 $\pm$ 0.3 & 90.4 $\pm$ 0.3 & \textbf{91.6 $\pm$ 0.4} & \textbf{92.3  $\pm$ 0.2}  \\
\textbf{en}-\textbf{ja} & 81.3 $\pm$ 0.1 & 86.7 $\pm$ 0.3 & \textbf{89.4 $\pm$ 0.2} & \textbf{89.1  $\pm$ 0.2} \\ \hline
\textbf{nl}-\textbf{en} & 72.3 $\pm$ 0.7 & 84.3 $\pm$ 0.1 & 84.0 $\pm$ 0.2 & \textbf{86.9 $\pm$ 0.1}  \\
\textbf{nl}-\textbf{fr} & 72.7 $\pm$ 0.1 & 81.5 $\pm$ 0.7 & 79.7 $\pm$ 0.1 & \textbf{83.2 $\pm$ 0.7} \\
\textbf{nl}-\textbf{es} & 69.2 $\pm$ 0.1 & 80.5 $\pm$ 0.4 & 79.3 $\pm$ 0.3 & \textbf{81.8 $\pm$ 0.4} \\
\textbf{nl}-\textbf{cs} & 50.0 $\pm$ 0.0 & 83.4 $\pm$ 0.5 & 50.0 $\pm$ 0.0 & 50.0 $\pm$ 0.0 \\
\textbf{nl}-\textbf{ja} & 72.2 $\pm$ 0.4 & 80.0 $\pm$ 0.7 & \textbf{81.9 $\pm$ 0.6} & \textbf{81.6 $\pm$ 1.1} \\
\hline
\textbf{fr}-\textbf{nl} & 75.3 $\pm$ 0.2 & 84.7 $\pm$ 0.3 & 80.6 $\pm$ 0.1 & \textbf{85.6 $\pm$ 0.8} \\
\textbf{fr}-\textbf{en} & 74.5 $\pm$ 0.4 & 87.7 $\pm$ 0.1 & \textbf{87.6 $\pm$ 0.0} & \textbf{89.1 $\pm$ 0.2} \\
\textbf{fr}-\textbf{es} & 72.1 $\pm$ 0.3 & 85.3 $\pm$ 0.2 & 82.0 $\pm$ 0.2 & 83.7 $\pm$ 0.4\\
\textbf{fr}-\textbf{cs} & 82.0 $\pm$ 0.8 & 87.2 $\pm$ 0.3 & 86.3 $\pm$ 0.4 & \textbf{86.9 $\pm$ 0.4} \\
\textbf{fr}-\textbf{ja} & 76.7 $\pm$ 0.6 & 81.4 $\pm$ 0.3 & \textbf{83.7 $\pm$ 0.5} & \textbf{84.0 $\pm$ 0.4} \\
\hline
\textbf{es}-\textbf{nl} & 78.2 $\pm$ 1.0 & 85.9 $\pm$ 0.6 & 82.7 $\pm$ 0.3 & \textbf{88.4 $\pm$ 0.4} \\
\textbf{es}-\textbf{fr} & 78.5 $\pm$ 0.4 & 87.5 $\pm$ 0.3 & 83.4 $\pm$ 0.2 & \textbf{88.3 $\pm$ 0.5} \\
\textbf{es}-\textbf{en} & 76.0 $\pm$ 0.1 & 88.8 $\pm$ 0.3 & 88.2 $\pm$ 0.2 & \textbf{90.6 $\pm$ 0.5} \\
\textbf{es}-\textbf{cs} & 82.5 $\pm$ 0.9 & 88.0 $\pm$ 0.4 & 87.0 $\pm$ 0.1 & \textbf{88.6 $\pm$ 0.3} \\
\textbf{es}-\textbf{ja} & 77.8 $\pm$ 0.7 & 83.1 $\pm$ 0.6 & \textbf{84.5 $\pm$ 0.7} & \textbf{84.7 $\pm$ 1.0} \\
\hline
\textbf{cs}-\textbf{nl} & 50.0 $\pm$ 0.0 & 81.1 $\pm$ 1.2 & 50.0 $\pm$ 0.0 & 50.0 $\pm$ 0.0 \\
\textbf{cs}-\textbf{fr} & 75.4 $\pm$ 0.5 & 81.4 $\pm$ 0.3 & 79.4 $\pm$ 0.7 & \textbf{86.4 $\pm$ 0.4} \\
\textbf{cs}-\textbf{es} & 72.6 $\pm$ 0.6 & 81.6 $\pm$ 0.9 & 79.1 $\pm$ 0.7 & \textbf{84.4 $\pm$ 0.4} \\
\textbf{cs}-\textbf{en} & 75.6 $\pm$ 0.5 & 84.5 $\pm$ 0.4 & \textbf{84.9 $\pm$ 0.4} & \textbf{88.2 $\pm$ 0.3} \\
\textbf{cs}-\textbf{ja} & 77.1 $\pm$ 0.9 & 80.5 $\pm$ 0.5 & \textbf{83.4 $\pm$ 0.7} & \textbf{84.5 $\pm$ 0.7} \\
\hline
\textbf{ja}-\textbf{nl} & 68.2 $\pm$ 0.7 & 76.9 $\pm$ 0.3 & 75.7 $\pm$ 0.3 & \textbf{81.9 $\pm$ 0.6} \\
\textbf{ja}-\textbf{fr} & 71.9 $\pm$ 0.1 & 77.8 $\pm$ 0.1 & 76.7 $\pm$ 0.5 & \textbf{83.2 $\pm$ 0.6} \\
\textbf{ja}-\textbf{es} & 68.9 $\pm$ 0.4 & 78.8 $\pm$ 0.5 & 76.1 $\pm$ 0.5 & \textbf{81.1 $\pm$ 0.3} \\
\textbf{ja}-\textbf{cs} & 77.9 $\pm$ 0.8 & 80.7 $\pm$ 0.4 & \textbf{82.2 $\pm$ 0.8} & \textbf{86.2 $\pm$ 0.9} \\
\textbf{ja}-\textbf{en} & 70.4 $\pm$ 0.4 & 81.6 $\pm$ 0.8 & \textbf{82.3 $\pm$ 0.5} & \textbf{84.4 $\pm$ 0.7} \\
\hline
\end{tabular}
\caption{\label{results3}
Macro-AUC scores (in \%) with vectors learned by leveraging aligned monolingual ontologies.
The first column shows the baseline results by relying on the aligned ontologies only.
The second column shows the results obtained by retrofitting \textbf{FT}$_{sif}$ embeddings to monolingual ontologies.
The next two columns show the results obtained by retrofitting \textbf{FT}$_{sif}$ embeddings to aligned monolingual ontologies when only the source embeddings or the target embeddings are known before retrofitting. The results larger than those in the column \textbf{Rfit$_{u\Omega}$}\textbf{FT}$_{sif}$ are in bold.
}
\end{small}
\end{table}

\begin{table}
\begin{small}
    \centering
    \begin{tabular}{|l|l|}
    \hline
\textbf{Embedding Model}    & \textbf{Dimension} \\ \hline
\textbf{FT} & 300 \\ 
\textbf{mBERT} & 768 \\
\textbf{XLM} & 2048 \\ 
\textbf{LASER} & 1024 \\ \hline
    \end{tabular}
    \caption{Embedding dimensions for pre-trained models used in our study, corresponding to the values provided and optimized by each model's authors.}
    \label{tab:my_label}
    
\end{small}
\end{table}

\end{document}